# Surveying You Only Look Once (YOLO) Multispectral Object Detection Advancements, Applications And Challenges

James E Gallagher[1], Edward J Oughton[1]
[1]Department of Geography & Geoinformation Science Department, George Mason University, Fairfax, VA 22030, USA.
Corresponding author: Edward J. Oughton (e-mail: eoughton@gmu.edu).

This work was supported by the Geography & Geoinformation Science Department at George Mason University.

**Abstract:** Multispectral imaging and deep learning have emerged as powerful tools supporting diverse use cases from autonomous vehicles, to agriculture, infrastructure monitoring and environmental assessment. The combination of these technologies has led to significant advancements in object detection, classification, and segmentation tasks in the non-visible light spectrum. This paper considers 400 total papers, reviewing 200 in detail to provide an authoritative meta-review of multispectral imaging technologies, deep learning models, and their applications, considering the evolution and adaptation of You Only Look Once (YOLO) methods. Ground-based collection is the most prevalent approach, totaling 63% of the papers reviewed, although uncrewed aerial systems (UAS) for YOLO-multispectral applications have doubled since 2020. The most prevalent sensor fusion is Red-Green-Blue (RGB) with Long-Wave Infrared (LWIR), comprising 39% of the literature. YOLOv5 remains the most used variant for adaption to multispectral applications, consisting of 33% of all modified YOLO models reviewed. 58% of multispectral-YOLO research is being conducted in China, with broadly similar research quality to other countries (with a mean journal impact factor of 4.45 versus 4.36 for papers not originating from Chinese institutions). Future research needs to focus on (i) developing adaptive YOLO architectures capable of handling diverse spectral inputs that do not require extensive architectural modifications, (ii) exploring methods to generate large synthetic multispectral datasets, (iii) advancing multispectral YOLO transfer learning techniques to address dataset scarcity, and (iv) innovating fusion research with other sensor types beyond RGB and LWIR.

## I. Introduction

Applying multispectral sensors to deep learning algorithms has emerged as a powerful tool for a myriad of applications [1], [2], [3]. Multispectral sensors capture data in both the visible and non-visible spectrum, providing rich information about the environment [4], [5], [6], [7], [8]. These sensors, combined with deep learning algorithms, such as computer vision and object detection models, have opened the door for task automation in the non-visible spectrum [9], [10], [11]. This field is leading to significant advancements that are solving real-world problems and revolutionizing industries [12], [13], [14]. As the development of affordable and compact multispectral sensors and embedded deep learning systems progresses, it has the potential to make these technologies more widely accessible, allowing for future large-scale implementation in resource-limited environments [15], [16].

One of the most popular, fastest, and easy-to-deploy open-sourced object detection algorithms is You Only Look Once (YOLO) [17], [18], [19]. Since its inception in 2015, YOLO has advanced computer vision incrementally with new YOLO releases [20]. As the number of YOLO variants has grown, so has the number of multispectral YOLO adaptations and applications [21], [22], [23]. Since default YOLO algorithms are optimized for Red-Green-Blue (RGB) images, there have been a plethora of YOLO multispectral variant modifications to enhance detection and classification performance, especially among humans and vehicle object classes [24], [25], [26], [27], [28]. As the democratization of multispectral object detection continues, many challenges remain [29], [30]. These include the need for more extensive and diverse multispectral datasets for model training and evaluation and multispectral model interpretability with hardware as edge device varieties increase at more affordable costs [31], [32], [33].

In its inception, the literature on multispectral sensors began with non-modified YOLO models being trained with multispectral imagery, providing baseline comparative results of using multispectral sensors for object detection [34], [35], [36]. As YOLO matured, the methods to carry out and synthesize multispectral feature extraction also advanced [37], [38], [39], [40], [41]. Various multispectral fusion methods were added within the backbone, neck, and head of the YOLO architecture, such as creating dual-stream learning approaches for fusing infrared and visible images to enhance detection performance [42], [43]. Multispectral YOLO applications have also been growing in various industries, including the agricultural sector, where specific vegetation can be detected, such as citrus trees or weeds in agricultural areas [44], [45], [46], [47]. Other highly cited examples of multispectral YOLO is the use of synthetic aperture radar (SAR) and other aerial-derived imagery to detect and track ships by optimizing multispectral YOLO models capable of sensing small objects [48], [49], [50].

This survey of the literature aims to provide a comprehensive overview of the evolution of multispectral object detection from 2020 to 2024, focusing on analyzing the use and adaptation of the YOLO convolutional neural network (CNN). This review will also identify growing trends related to remote sensing applications combined with YOLO neural network adaptations to optimize detection performance along the non-visible spectrum. The research questions include:

1. How have modifications and enhancements to the YOLO architecture impacted its performance and adaptability for multispectral imaging applications compared to default YOLO models?
2. What sensors, collection platforms, and object classes are growing in use-case applications with YOLO-based multispectral object detection?
3. What are the main challenges and future research directions for YOLO-based multispectral object detection?

This survey will follow the following structure to organize the literature thematically and to assist in answering the research questions. Section II will discuss the methods for surveying the literature. Section III will discuss the history and evolution of YOLO, while Section IV will summarize and quantify multispectral imaging technologies and platforms. Section V will analyze cross-cutting themes between YOLO and multispectral. Section VI will discuss YOLO adaptations for multispectral object detection. The survey will then discuss datasets and evaluation metrics used in multispectral YOLO applications in Section VII. Section VIII will return to the discussion of YOLO multispectral object detection, the challenges in the field, and its future direction. Finally, Section IX will provide concluding remarks.

**II. Method For The Survey Of The Literature**

In this rapidly evolving field, this review comprehensively summarizes the recent advancements in YOLO-based multispectral object detection from January 2020 to April 2024. We evaluated 400 papers from reputable journals that meet the pre-specified conditions in concordance with the research question. A search methodology for identifying, selecting, and analyzing relevant literature was designed to ensure a thorough and systematic approach. In the end, we selected 200 papers from reputable journals that were cutting-edge and highly cited. Additionally, several novel non-YOLO CNN object detection research papers were retained to compare multispectral computer vision development outside of YOLO.

The search strategy focused on publications within the specified timeframe, utilizing major academic databases, including IEEE Xplore, Scopus, Web of Science, and Google Scholar. The primary keywords used in the search were YOLO, CNN, object detection, multispectral, MSI, HSI, CNN, LWIR, and thermal. These keywords were combined to refine search results while maximizing relevant results.

Only peer-reviewed journal articles, conference proceedings, and preprints from reputable sources were included in this review. After 400 research papers had been collected, a package was installed into Zotero to filter papers by citation count. The papers with the greatest relevance and highest citation counts were selected, ensuring the inclusion of high-quality literature. Following this step, 200 papers were retained for analysis. However, citation count was not the ultimate deciding factor. A select number of recently published papers from 2023-2024 with novel approaches were kept despite having a lower citation count.

The selected papers were aggregated into a spreadsheet, where key metrics were extracted to analyze publication trends in the literature. We extracted information from each selected paper, including publication details, YOLO variant, multispectral sensors, platforms employed, object classes detected, dataset type, performance metrics, key findings, and contributions. A Python notebook was also built to extract keywords from the literature to create an n-gram plot (Fig. 4B). This systematic data extraction allowed for a comprehensive analysis and comparison across studies, ensuring a thorough review. Selected papers were analyzed based on methods and use cases, such as the type of multispectral sensor used, application domain, YOLO architecture modifications, and performance improvements over previous methods. This categorization facilitated the identification of trends and patterns in the field. Particular attention was paid to innovative approaches that addressed limitations in existing multispectral object detection methods that opened new avenues for research.

In summary, 200 papers that met the evaluation criteria were reviewed, providing a comprehensive overview of YOLO-based multispectral object detection advancements. This systematic approach allows for synthesizing the current knowledge in the field, identifying key trends and challenges, and highlighting promising directions for future research.

**III. The History Of YOLO**

**A. YOLOv1**
Joseph Redmon revolutionized computer vision when he released YOLOv1 in 2015 [51]. Unlike previous CNN architectures, like AlexNET, which only did image classification, or R-CNN, which used a slow two-stage detection method, YOLO treats object detection as a single-pass regression problem, making it both fast and computationally efficient [52], [53]. This single-pass approach made real-time object detection a reality. What separated YOLO apart from other CNNs is the ability to use single-pass regression to place both bounding boxes and class probabilities based on image regions [54]. This approach eliminates traditional CNN processes, such as separate region proposal steps, resulting in YOLO performing significantly faster than the computationally inefficient two-stage detectors such as R-CNN, Fast R-CNN, and Faster R-CNN [52], [54], [55]. Despite its novel concept to improve inference speed, YOLOv1 remained less accurate than the leading Fast R-CNN models. Using the VOC 2007 dataset, YOLOv1 achieved an mAP of 64.4% at 45 frames per second (FPS), with Fast R-CNN achieving 70% mAP at 0.5 FPS [56].

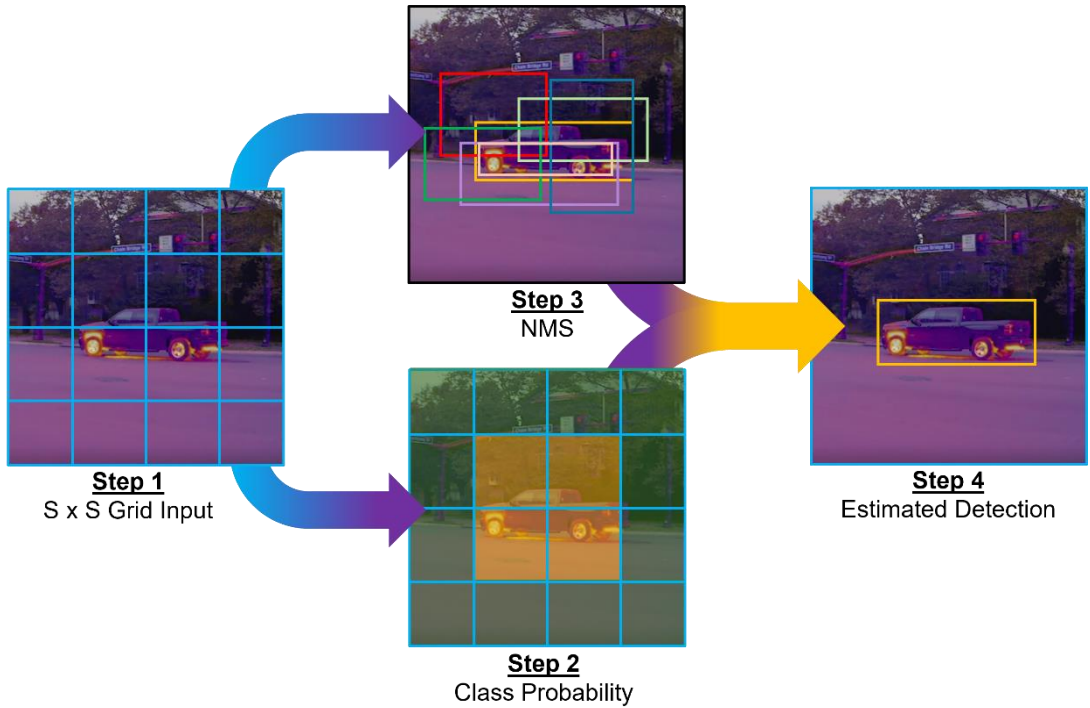

**FIGURE 1.** A simplified figure on how YOLO conducts object detection and classification on a fused RGB-LWIR image in real-time.

Figure 1 illustrates how YOLO conducts object detection through a single pass of a neural network. The process begins with grid segmentation, where the input image is divided into an S × S grid (7 × 7) [57]. Each grid cell predicts bounding box location concurrently with associated confidence scores and class probabilities for detected objects. Each bounding box prediction comprises five elements (x, y, w, h, and confidence) [56], [58], [59]. The x and y coordinates are of the box's center relative to the grid cell, while the width and height is the box's relative position to the entire image. Lastly, the confidence score reflects the model's certainty of object class presence and localization accuracy [56], [57].

The network yields a unified output in the form of a tensor with dimensions S × S × (B × 5 + C), followed by Non-Maximum Suppression (NMS) to remove redundant detections [57], [60]. As depicted in Figure 1, NMS is applied to eliminate redundant detections, refining the output to the most probable and accurate bounding boxes. This regression-based output allows YOLO to conduct simple and fast detection and classification, counter to the traditional complex models such as Fast R-CNN that employ separate classification and regression outputs [61]. Despite performing faster than traditional CNNs, the accuracy is still not as high when compared to two-stage detectors [52].

**B. YOLOv2**
Shortly after the release of YOLOv1, Redmon released YOLOv2, which built upon YOLOv1's architecture and features the darknet-19 framework, consisting of 19 convolutional layers and five max pooling layers [17], [62]. In his paper, Redmon states that YOLOv2 achieved 76.8% mAP on the VOC 2007 dataset at 67 FPS outperforming R-CNN, ResNet, and SSD [17]. YOLOv2 introduced several improvements, which include batch normalization to address internal covariate shift, a high-resolution classifier for improved classification performance, convolutions with anchor boxes for increased recall, and many other improvements [17], [62], [63].

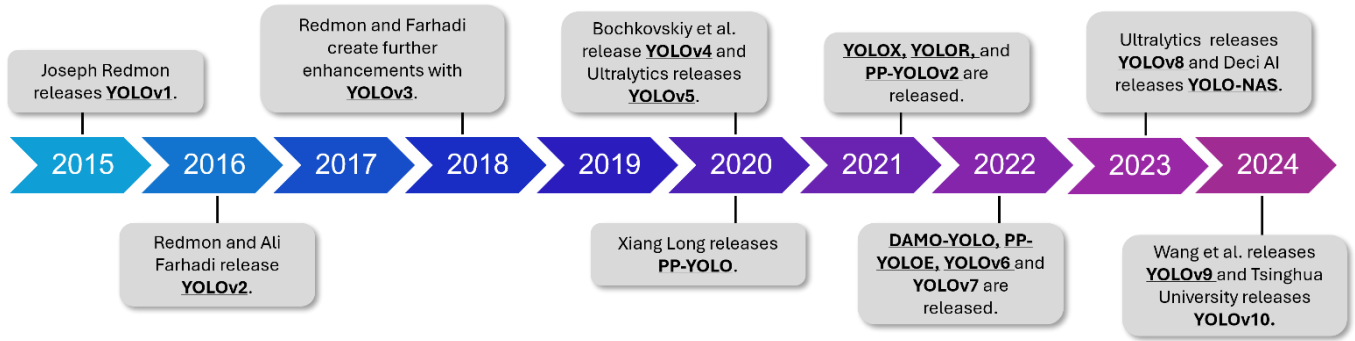

FIGURE 2. A timeline depicting the releases of significant YOLO models from 2015 until July 2024.

### C. YOLOv3

Redmon and Farhadi released their last version of YOLO, YOLOv3, in 2018 [64]. The primary advancement that YOLOv3 had over YOLOv2 is that it used a deeper feature extractor called Darknet-53, which was a substantial improvement over the previously used Darknet-19 in YOLOv2 [64], [65]. Furthermore, YOLOv3 predicted bounding boxes at three different scales, a principle similar to feature pyramid networks, to help improve the detection of objects of various sizes [66]. YOLOv3 also employed dimensional clusters as anchor boxes, like YOLOv2, predicting three boxes in each scale [67]. This resulted in nine anchor boxes [68]. Combined, these enhancements improved detection accuracy in YOLOv3 when compared to YOLOv2 by a staggering 13.9% mAP when applied to the COCO dataset [64], [69], [70].

### D. YOLOv4, YOLOv5, and PP-YOLO

YOLOv4 and YOLOv5 were both released in 2020. YOLOv4, released by Bochovskiy et al., used CSPDarknet-53 as the backbone and added Spatial Pyramid Polling (SPP) and Path Aggregation Network (PAN) in the neck [71]. Another improvement in YOLOv4 was the introduction of the mosaic data augmentation technique, which combined four training images into one. This technique improves the detection of small objects [71], [72]. YOLOv5, released by Glenn Jocher, was a paradigm shift from previous YOLO versions because of its transition away from Darknet and towards PyTorch, leading to an increase in accessibility and ease of use for developers [62], [73]. YOLOv5 also implemented a Spatial Pyramid Pooling Fast (SPPF) layer, replacing the SPP layer used in YOLOv4 [74]. This allowed for pooling features of different scales into a fixed feature map, increasing inference speed [75]. Furthermore, YOLOv5 added additional image augmentation techniques, including random affine, MixUp, mosaic, and HSV augmentation [62], [76]. The last YOLO variant released in 2020 was PP-YOLO by Long et al. [77]. PP-YOLO was a modification of YOLOv3 and uses the PaddlePaddle framework, differing from YOLOv5's PyTorch framework [78], [79]. The goal of PP-YOLO was to balance effectiveness and efficiency in the CNN for practical, real-world applications [77]. PP-YOLO uses ResNet50-vd-dcn as the backbone, while YOLOv5 uses CSP [80], [81]. PP-YOLO also uses different optimization techniques that differ from YOLOv5, which include IoU loss, IoU Aware, grid sensitivity, and SPP [82], [80]. Conversely, YOLOv5 uses techniques like Mosaic Augmentation, adaptive anchor leaning, and image scaling for model optimization [83], [84].

### E. YOLOR, YOLOX, and PP-YOLOV2

In 2021, YOLOR, YOLOX, and PP-YOLOv2 were released. YOLOR (You Only Learn One Representation) was designed for multitask learning [85], [86], [87]. YOLOR leverages explicit and implicit knowledge or data observations and learned latents, allowing the model to capture and share semantic information across tasks [85], [88]. While previous YOLO models were designed for object detection, YOLOR is multitask-focused and designed for object detection, instance segmentation, image captioning, and semantic segmentation [85], [89].

YOLOX was released in 2021 by Ge et al. [90]. Unlike YOLOv5, which used a coupled head, YOLOX used a decoupled head for classification and localization, improving speed and performance [91]. Furthermore, unlike YOLOv5, YOLOX uses an anchor-free design, simplifying detection and reducing parameters [62], [92]. YOLOX also uses a non-traditional label assignment called SimOTA, allowing for a more dynamic and effective labeling strategy than YOLOv5's traditional label strategy [93]. SimOTA reduces training time and increases detector performance from 45.0% AP to 47.3% AP [90]. The last YOLO variant released in 2021 was PP-YOLOv2 by Huang et al. [94]. PP-YOLOv2 builds on PP-YOLO with additional modifications to increase performance and inference speed [95]. Using the COCO dataset, PP-YOLO achieved an mAP of 45.9% at 72.9 FPS, while PP-YOLOv2 achieved an mAP of 49.5% at 68.9 FPS [94]. PP-YOLOv2 has several modifications from the original version. Firstly, PP-YOLOv2 adds a Path Aggregation Network (PAN), enhancing feature fusion across various scales [96]. Mish Activation Function was also added to PP-YOLOv2 [96], which later proved effective in YOLOv4 and YOLOv5 [97], [98]. However, unlike YOLOv4 and YOLOv5, PP-YOLOv2 integrates the Mish Activation function into the neck, keeping the backbone unchanged [94], [99].

### F. YOLOv6, YOLOv7, DAMO-YOLO and PP-YOLOE

In 2022, YOLOv6, YOLOv7, DAMO-YOLO, and PP-YOLOE were introduced. YOLOv6 was released by Li et al. and represented a significant redesign of the YOLO architecture from YOLOv5 [100]. YOLOv6's backbone (EfficientRep), neck (Rep-PAN), and head (decoupled head) were all changed from YOLOv5's backbone (CSP), neck (CSP) and head (coupled head) [100], [101], [102]. YOLOv6 also utilized a different loss function (VariFocal Loss) and label assignment called Task Alignment Learning (TAL) than YOLOv5's SimOTA [103]. YOLOv7 was released by Wang et al. and had several modifications from YOLOv6 [104]. While YOLOv6 emphasized quantization and model deployment for industrial applications, YOLOv7 focused more on architectural innovations and training strategies to increase performance [105]. The maximum AP of YOLOv7-E6E (56.8%) was 13.7% higher than YOLOv6 (43.1%) on the COCO dataset [104]. Unlike YOLOv6, which uses EfficientRep and CSPStackRep architecture, YOLOv7 introduced the Extended Efficient Layer Aggregation Network (E-ELAN) architecture [106], [107].

DAMO-YOLO was also released in 2022 by Xu et al. and introduced a myriad of modifications [108]. While YOLOv7 used the E-ELAN backbone and neck and YOLOv6 using a CSPStackREP backbone, DAMO-YOLO implements a MAE-NAS backbone [21], [108]. When analyzing performance, DAMO-YOLO-M achieved an AP of 67.2%, while YOLOv6-M achieved an AP of 66.8%, with YOLOv5-M achieving 64.1% AP [108]. Despite the AP being very similar between these three models, DAMO-YOLO has fewer parameters than YOLOv6-M, decreasing latency by up to .10 ms [108]. Lastly, PP-YOLOE was released by Xu et al. with the intent of building on the success of PP-YOLOv2, which was used primarily for industrial applications [109]. PP-YOLOE is an anchor-free algorithm that is compatible with different hardware, making it easy to deploy when compared to other YOLO variants [110]. Unlike PP-YOLOv2, which used an anchor-based detection approach, PP-YOLOE used an anchor-free approach [109], [111]. PP-YOLOE also used a different backbone, neck, and head from PP-YOLOv2, performing 1.9% mAP higher than PP-YOLOv2 while also increasing inference speed by 13.35% [109].

### G. YOLOv8, YOLOv9, AND YOLOv10

YOLOv8 was released in 2023 by Ultralytics, the company that also created YOLOv5 [112]. According to Glenn Jocher, the founder of Ultralytics, YOLOv8 is an improved version of YOLOv5, using a similar backbone with significant adjustments to the CSPLayer, now referred to as the C2fmodule in YOLOv8 [62], [113]. While YOLOv7 focused on optimizing object detection performance, YOLOv8 focused on semantic segmentation capabilities [114]. When tested against the COCO dataset, YOLOv8 achieved an mAP of 53.9%, while YOLOv7 achieved an mAP of 51.2% [62]. Also released in 2023, Deci developed YOLO-NAS, which was designed to improve the detection of small objects while optimizing the performance-per-compute ratio [115], [116], [117]. This allowed for easier deployment on edge devices and enhanced localization accuracy [57]. Unlike previous YOLO models, which were designed and built with manual human effort, YOLO-NAS's architecture was created using AutoNAC, a proprietary neural network created by Desi [57], [118]. YOLO-NAS was also optimized for ease of deployment on edge devices by adapting the architecture to hardware constraints [119].

YOLOv9 was released in 2024 by Wang et al., who was the same creator of YOLOv7 [120]. YOLOv9 introduced a new architecture called Generalized Efficient Layer Aggregation Network (GELAN) [121]. The primary advancement in YOLOv9 from YOLOv7 and YOLOv8 addressed data loss during transmission through the neural network [122]. YOLOv9 also increased its ease of use and efficiency compared to previous YOLO models due to fewer parameters, resulting in a computationally less intensive model than previous YOLO models [120], [123]. Compared to YOLOv8-X, YOLOv9-E has 16% fewer parameters, resulting in a 27% decrease in computations while improving mAP by 1.7% [120].

Finally, YOLOv10 was released in May of 2024 by Wang et al. from Tsinghua University [124]. The goal of YOLOv10 was to decrease the reliance on NMS for post-processing, which impedes the deployment of YOLO for real-world applications due to latency issues [124], [125]. Not using NMS makes YOLOv10 easier to deploy with faster inference speed for real-world applications [126]. YOLOv10-M matches YOLO-9-M's mAP (51.1%) while having 23% fewer parameters. YOLOv10 outperforms all YOLOv8 variants by 1.2-1.4% mAP with a 28-57% parameter decrease [25]. When comparing model performance between YOLOv9 and YOLOv10, both YOLO versions have similar mAP performance. However, because YOLOv10 does not use NMS, it is easier to deploy than YOLOv9 due to fewer parameters.

### IV. Multispectral Sensors And Platforms

Multispectral sensor usage is growing in the field of computer vision because of the ability to capture visible and non-visible data across the electromagnetic spectrum [127], [128], [129]. Fig. 3 visualizes the electromagnetic spectrum, with spectral bands and wavelength ranges in nanometers (nm) on the bottom of the figure, while the top of the figure lists the sensors that align with their respective spectral bands. Beginning on the left side, the first spectrum is ultraviolet, with a spectral band ranging from 100 to 400 nm [130]. In this survey, only one paper used an ultraviolet sensor to conduct object detection on solar activity on the sun's surface [131]. The next spectral band is the visible spectrum (RGB), which has a spectral range of 450 nm (blue) to 750 nm (red) [132]. In this survey, 109 studies used RGB combined with other sensors that operate along the infrared spectrum. RGB is a popular sensor that can be combined with other non-visible spectrums because it is the most widely used and available sensor that provides high levels of image details at a low cost [133], [134].

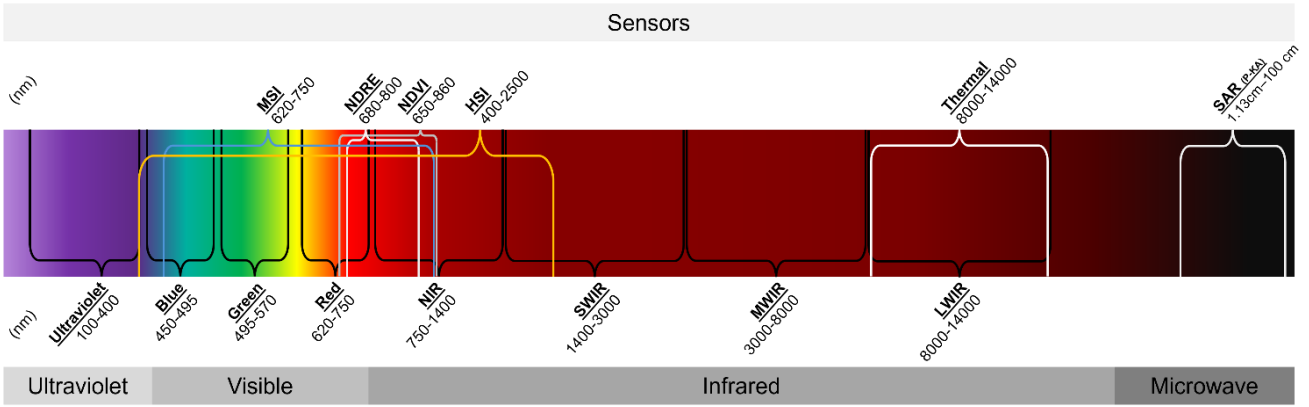

**FIGURE 3.** The electromagnetic spectrum broken down my sensor type with their corresponding spectrum range in nanometers (nm).

After RGB is the Near Infrared (NIR) spectrum capable of collecting data between 750 to 1400 nm [135]. In this survey, 12 studies used NIR or an RGB-NIR fusion approach for object detection. NIR is a reliable band along the infrared spectrum because of its ability to work in poor-visibility conditions while maintaining image resolution at long distances [136], [137], [138]. After NIR is the Short-Wave Infrared (SWIR) band, operating between 1400 and 3000 nm [139]. Only one study in this review used fused RGB-SWIR object detection [140]. After SWIR is Medium-Wave Infrared (MWIR), ranging from 3000 to 8000 nm [141]. This survey did not include any use-case examples using MWIR. At the end of the infrared spectrum is Long-Wave Infrared (LWIR), which has a spectral range of 8000 to 14000 nm [142]. Sixty-three studies in this survey used a combination of LWIR or fused RGB-LWIR for object detection applications. Finally, five research papers in the survey used SAR, with three of the five papers using modified YOLOv5 models to more accurately identify small ships and ship orientation in satellite-based SAR imagery [48], [143], [144].

### A. Multispectral Imaging (MSI)
When looking at the sensor list at the top of Fig. 1, Multispectral Imaging (MSI) is the first sensor operating within the RGB and NIR bands. MSI sensors collect data along the blue, green, red, red edge, and infrared bands [145].

Red edge is a narrow band between red and NIR (centered between 700-705 nm) frequently used to determine plant health and to calculate the Normalized Difference Red Edge (NDRE) index [146], [147], [148]. Conversely, the Normalized Difference Vegetation (NDVI) index is a combination of the red band and NIR band to frequently calculate vegetation wellness (650-860 nm) [6]. MSI is used primarily to determine vegetation health and is used widely in the agro-food industry [149]. Although the number of spectral bands an MSI sensor can collect depends on camera specifications, space-based MSI typically consists of thirteen bands [150].

### B. Hyperspectral Imaging (HSI)
The next sensor type is Hyperspectral Imaging (HSI). In this survey, only 19 studies used HSI for object detection. HSI has a spectral range of 400-2500 nm, allowing for the analysis of hundreds of bands along the RGB, NIR, and SWIR wavelengths [151]. Like MSI, the number of bands HSI can collect depends on the sensor specification. In this survey, the HSI sensor with the highest number of bands collected was from the Moffett Field Dataset with 224 bands ranging from 400- to 2500 nm [152].

### C. Long-Wave Infrared (LWIR)
The next and most common sensor type used in the non-visible spectrum is LWIR. Other names for LWIR are Thermal Infrared (TIR) or Forward Looking Infrared (FLIR) [153]. LWIR is the most widely used sensor for multispectral object detection for a variety of reasons. Firstly, LWIR can be used in real-time, allowing for several practical applications like autonomous driving [154], [155], [156]. Secondly, LWIR sensors provide thermal data that can be used as additional unique edges for object detection algorithms to train on [34]. The third reason is that LWIR sensors work in complex visibility conditions [157], [158]. The cost of LWIR sensors has also steadily decreased, making them increasingly available. Lastly, fusing RGB with LWIR provides edge redundancy and edge enhancement for CNN models to train on, thus improving their performance and increasing model resiliency to changing illumination conditions [159].

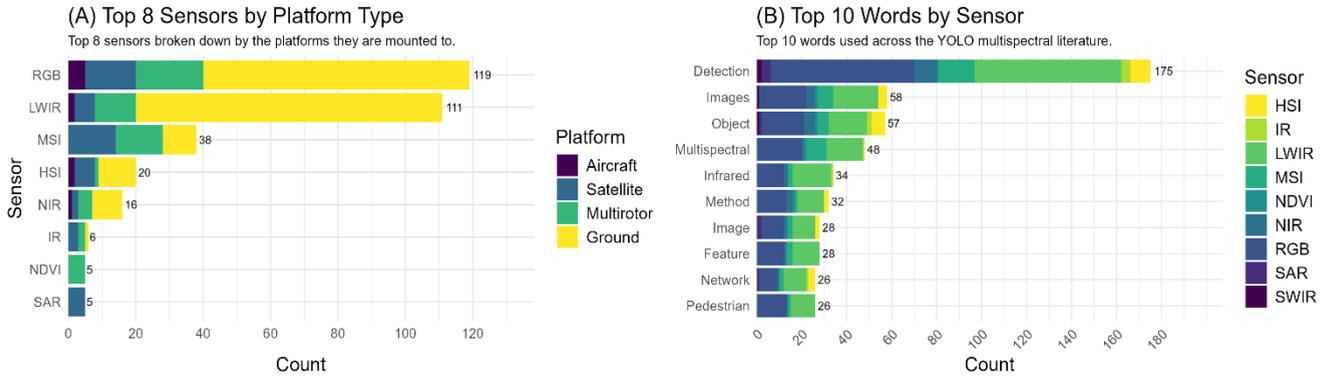

**FIGURE 4.** Quantifying sensor types, vehicle platforms, and most common words used in this survey.

**D. Vehicles Used For Sensor Deployment**
Multispectral sensors can be deployed from various platforms, such as ground-based systems, multirotors, aircraft, and satellites, each with advantages and limitations. For example, ground-based sensors offer high spatial resolution but limited coverage [160]. Aerial platforms, such as drones and aircraft, balance coverage and resolution, making them suitable for precision agriculture and infrastructure monitoring applications [161], [162]. Satellite platforms offer extensive coverage at the cost of lower spatial resolution, making them useful for large-scale environmental monitoring and land-use mapping [163]. The desired selection of multispectral imaging technology and platform depends on the application's requirements. For example, precision tasks, such as agriculture or autonomous driving, benefit from high-resolution ground-based sensors, while environmental monitoring may rely on satellite-based multispectral data [164], [165], [166].

**E. Ground-Based Sensors**
In this survey, ground-based sensors were the most used approach for data collection, with 127 studies deploying ground-based sensors (Fig. 4A). Within ground-based sensors, RGB sensors were the most widely used (78 studies), followed by LWIR (91 studies). Ground-based RGB and LWIR applications centered around pedestrian detection, with 59 studies on LWIR human detection and 50 on RGB human detection (Fig. 7F). Additionally, there were 48 studies on fused RGB-LWIR ground-based object detection. Although LWIR sensors have many advantages, they have low-resolution and basic color features [167]. Because of this, RGB and LWIR are commonly used to enhance feature extraction for object detection in environments with challenging visibility conditions. There have been several YOLO modifications that optimize the detection of low-resolution thermal objects in ground-based imagery [168].

These modifications are primarily for pedestrian and vehicle detection in complex illumination conditions [169], [170], [171]. Ground-based MSI and HSI were used primarily for agricultural and industrial applications, such as detecting coal gangues, organisms of interest in agricultural products, and tracking vehicles [172], [149], [173], [174].

**F. Multirotor Drones**
The next most common platform used in multispectral object detection are multirotor drones, with 36 studies using this approach. The primary benefit of multirotors is that they are low-cost, easy to deploy, and can be easily modulated with various sensors required for the task. MSI is the most used sensor type deployed on multirotors for multispectral object detection (20 studies). A typical application for drone-based MSI is for agricultural purposes to detect specific plant-based object classes, such as weeds, plant-borne diseases, and plant species identification [45], [175], [176]. The use of multirotors for multispectral object detection in this survey has doubled between 2020 and 2023, showing the growing use of this platform due to increased accessibility.

**G. Satellites**
Satellites were this survey's third most commonly used platform for multispectral object detection (29 studies). High-quality MSI data is openly available from sources such as the European Space Agency's Sentinel-2 and NASA's Landsat program, allowing for a myriad of multispectral object detection research and applications [177], [178]. The challenge with satellite-based multispectral object detection is low resolution. Because of this, object detection models are frequently modified to perform optimally when presented with low-resolution data. For example, LMO-YOLO is a modified YOLO CNN designed to detect ships in low-resolution satellite images [179].

**H. Aircraft**
Lastly, aircraft were the least used data collection platform in this survey (12 studies). The reason for their low use is the higher costs of collecting multispectral data when compared to open-sourced satellite data or low-cost multirotor methods. Similarly to satellite-derived data, images from aircraft also suffer from low resolution.

**I. Advancements And Challenges In Multispectral Sensors**

There is growing interest in developing low-cost multispectral sensors that can be miniaturized for deployment on unmanned ground and air systems. These sensors typically use compact and lightweight optics, such as micro-lens arrays and dichroic filters, to capture multispectral data with high spatial and spectral resolution [180]. Integrating these sensors with embedded deep learning algorithms can enable the real-time processing and analysis of multispectral data on the edge. When connected to cellular networks, such as a 5G network, the data range and processing speed of edge devices with multispectral sensors can be significantly increased with centralized computing [181].

Another critical development in multispectral object detection is using event-based sensors, which capture temporal changes in the scene. Such an application can be used to conduct real-time vegetation stress detection in the agricultural industry [182]. Integrating event-based sensors with deep learning algorithms can enable the development of novel solutions, such as identifying potential landslide areas in mountainous regions before they occur [183].

Despite the advances in multispectral imaging technologies and platforms, several multispectral sensor challenges are mentioned in the literature. One of the main challenges is the alignment, calibration, and registration of multispectral data captured by different sensors and platforms [184], [185]. Multispectral images can be affected by various factors, such as illumination, sensor noise, lens distortion, and atmospheric effects, which can degrade the quality and consistency of the data [186], [187]. To address this challenge, researchers have proposed various calibration and registration methods, such as geometric and radiometric calibration, image fusion, and point cloud registration, which can improve the accuracy and reliability of multispectral data [185], [188], [189]. Another challenge is the storage and management of large-scale multispectral data, which can involve significant computational cost and storage resources [190].

Lastly, when analyzing the most common words found in the survey (Fig. 4B), the word *detection* was the most frequent word for RGB, LWIR, and MSI sensors. The following three words, including "*images*", "*object*", "*multispectral*", were found almost equally. When analyzing who is conducting the predominance of multispectral object detection, Fig 5. visualizes country-model-sensor-type research, while Fig. 6A plots this data on a world map, while Fig. 6B plots the location of the primary author's home city in China. Both figures clearly illustrate that Chinese academic institutions dominate multispectral object detection research, with 116 of the 200 studies (58%) analyzed in this survey. Additionally, China is leading the research in modifying YOLO algorithms for multispectral applications (with 65 out of the 83 studies on multispectral YOLO).

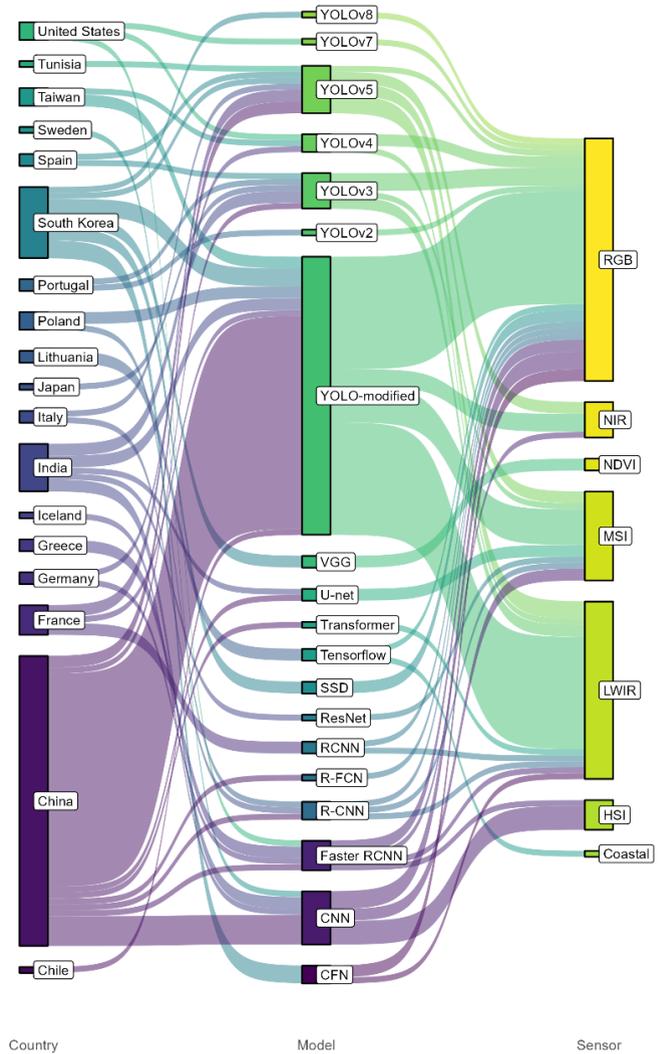

**FIGURE 5.** Visualizing Country, model, and sensor types analyzed in this survey.

**V. Cross-Cutting Themes**

The literature reveals several cross-cutting themes in YOLO-based multispectral object detection research from 2020 to 2024. These themes highlight the evolving approaches, challenges, and opportunities in this rapidly advancing field. The primary cross-cutting themes are architectural innovations, detecting small objects, domain-specific multispectral model adaptations, and model optimization for real-time applications.

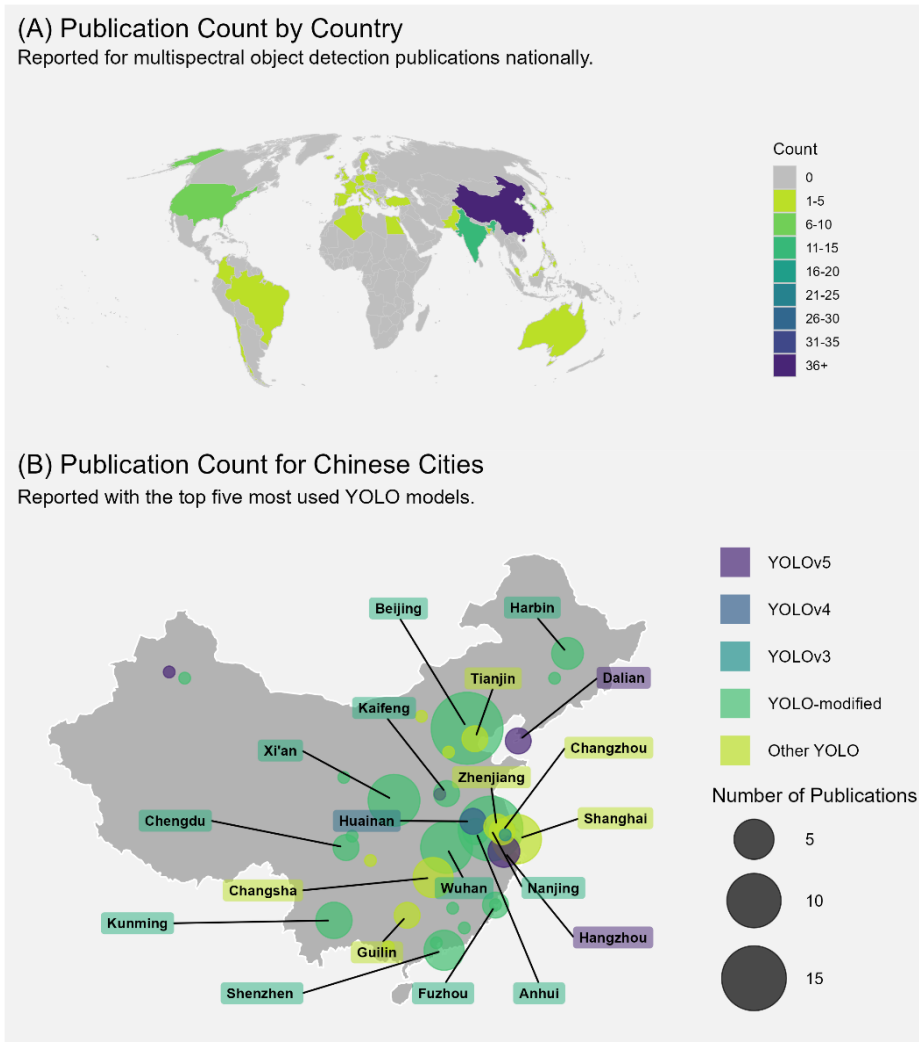

**FIGURE 6.** A geographic overview of where multispectral object detection research is being conducted.

### A. Architectural Innovations

A predominant theme across the literature is the development of architectural modifications from RGB-centric default YOLO models to YOLO models optimized for multispectral applications. Various architectural approaches have been developed to extract pertinent information from different spectral bands, particularly in the RGB and LWIR spectrum. One common approach is the adoption of dual-stream architectures, where separate branches in the neural network process different spectral inputs. For instance, Shao et al. proposed MOD-YOLO, which uses a dual-stream feature extraction network and a transformer-based cross-modal spectral multi-scale feature fusion module [169]. Similarly, Sun et al. introduced GMD-YOLO, featuring a dual-channel CSPDarknet53 backbone with a Ghost module for feature extraction [170]. These dual-stream approaches allow for specialized processing of each modality in the backbone prior to data integration into the neck.

Another recurring architectural innovation is the use of attention mechanisms for adaptive fusion. Chen et al. developed TF-YOLO, which incorporates a transformer-fusion module to integrate key features between visible and infrared images [171]. This approach allows the model to dynamically adapt to changing illumination conditions. Bao et al. proposed Dual-YOLO, which uses attention fusion and fusion shuffle modules to reduce redundant fusion feature information and enhance complementary characteristics of infrared and visible images [190].

These architectural innovations reflect a growing demand for specialized YOLO models that effectively handle real-time multispectral data in adaptive illumination conditions. These models can extract and synthesize critical information from different spectral bands by developing modality-specific processing streams and adaptive fusion mechanisms within the YOLO neural network.

## B. Small Object Detection

Accurate detection of small objects is a significant challenge in multispectral object detection, particularly in airborne remote sensing applications. This issue is exacerbated in low-resolution imagery, such as imagery obtained from satellite or high-altitude aerial platforms, where spectral resolution can become an issue. Several studies have proposed modifications to YOLO architectures to enhance small object detection capabilities.

Pham et al. introduced YOLO-Fine, a modified YOLO architecture specifically designed for detecting small objects in remotely sensed images [49]. Their approach included refinements to the detection grid and feature extraction process to better handle small targets. Similarly, Xu et al. developed LMO-YOLO for ship detection in low-resolution satellite imagery, incorporating multiple linear rescaling and dilated convolutions to enhance feature extraction for small targets [179]. Mou et al. proposed YOLO-FR, a YOLOv5-based algorithm that uses a feature reassembly sampling method to improve the detection of small infrared targets [168]. Zhou et al. introduced YOLO-SASE, which combines super-resolution techniques with a novel Self-Adaption Squeeze-and-Excitation (SASE) module to enhance small target detection in complex backgrounds [191].

These approaches demonstrate a common theme of adapting YOLO architectures to better handle the scale and resolution challenges inherent in multispectral imaging, particularly for small object classes. By incorporating techniques such as multi-scale feature fusion, super-resolution, and specialized attention mechanisms, the sensitivity and accuracy of YOLO models for small target detection across various spectral domains will continue to improve and advance.

## C. Domain-Specific Adaptations

The literature also reveals a trend towards adapting YOLO models for specific application domains, leveraging the unique characteristics of multispectral data in various conditions. This theme is most evident in agriculture, infrastructure monitoring, and defense applications.

Several studies have designed YOLO models that are enhanced for crop monitoring and pest detection using multispectral imagery in agriculture. Osco et al. developed an approach for counting and geolocating citrus trees in UAV multispectral imagery [44]. Osorio et al. used YOLO for weed detection in lettuce crops using multispectral images [45]. These studies demonstrate the potential of multispectral YOLO variants to address specific agricultural challenges, leveraging spectral information to improve crop management and pest control.

In the realm of plant disease detection, Georgantopoulos et al. created a multispectral dataset for early detection of tomato plant diseases, specifically targeting Tuta Absoluta and Leveillula Taurica, using deep learning techniques [192]. Rouš et al. applied YOLOv5 with multispectral imaging for detecting apple scab in orchards, highlighting the potential of combining deep learning with multispectral data for early disease detection [173]. Park et al. developed a multichannel CNN model for detecting pine wilt disease using drone-based multispectral imaging, achieving high performance in identifying diseased trees [193].

Integrating YOLO variants with multispectral imaging has also shown promise in weed and pest control applications. Naveed et al. proposed a saliency-based semantic model for weed detection and classification using multispectral imaging from UAVs, offering an unsupervised approach that does not require extensive training data [147]. Pansy and Murali utilized UAV hyperspectral imaging and advanced machine learning techniques for early detection and management of diseases and pests in mango crops, demonstrating the potential of high-dimensional spectral data in pest control [175]. For better agriculture planning and management, Shen et al. employed YOLOv8 with multispectral remote sensing for estimating maize planting densities [194]. Strzępek et al. used Detectron2 for object detection and segmentation in drone-based multispectral imagery, enabling comprehensive crop analysis [195].

Researchers have also adapted YOLO to optimize detections in thermal and multispectral imagery for infrastructure monitoring for various inspection tasks. Chen et al. proposed a method based on YOLOv5 and multiscale data augmentation for visual inspection in substations [196]. Lei et al. developed a Deeplab-YOLO method for detecting hot-spot defects in infrared images of PV panels [197]. These applications showcase the versatility of YOLO-based approaches in leveraging multispectral data for infrastructure maintenance and safety.

In defense and surveillance applications, several studies have focused on adapting YOLO models for thermal and multispectral imaging to enhance detection capabilities in complex and adaptive conditions. The integration of multispectral imaging, particularly the combination of RGB and LWIR sensors, has shown significant promise in improving object detection performance in various military and security applications. Kristo et al. (2020) explored thermal object detection using YOLO in challenging weather conditions, such as fog and rain, aimed at enhancing security systems through improved automatic human detection [34]. Their study demonstrated that default YOLOv3 retrained on thermal images significantly enhances the ability to detect humans in difficult weather conditions.

In the realm of camouflage detection, Wang et al. developed a deep learning-based multispectral method for detecting camouflaged people [198]. Their proposed MS-YOLO model achieved an mAP of 94.31% and was capable of real-time detection at 65 FPS, demonstrating high efficacy in detecting camouflaged individuals in various desert and forest scenes. This research has relevant military applications where rapidly detecting concealed targets is crucial to mission success. McIntosh introduced a novel network, TCRNet, optimized to detect targets in cluttered infrared imaging environments by enhancing the "target to clutter ratio" (TCR) [199]. This approach, developed for the U.S. Army, demonstrated a significant improvement in detection probability by more than 30% and reduced the false positive rate by more than a factor of 2 compared to leading methods like Faster R-CNN and YOLOv3 [199].

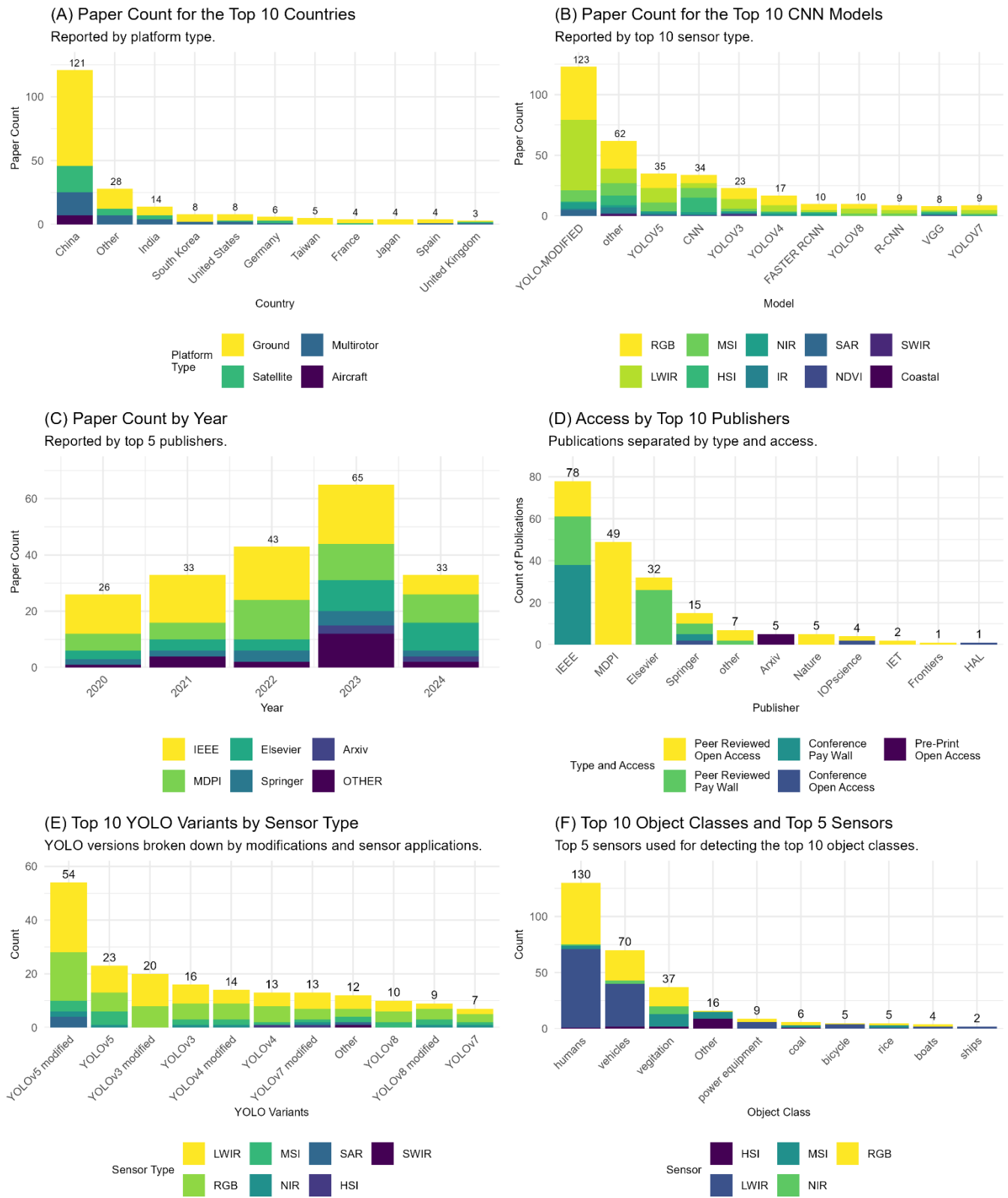

**FIGURE 7**. Panel plot of the key variables summarized in the survey.

Kwan and Gribben focused on improving target detection in long-range and low-quality infrared videos using deep learning techniques, specifically addressing the challenges posed by small target sizes and poor video quality [200]. Their work, also conducted for the U.S. Army, achieved a 95% detection accuracy of enemy vehicles, showcasing the potential of YOLO-based approaches in long-range surveillance and target acquisition.

The application of multispectral YOLO variants extends to maritime defense as well [201]. Sun et al. (2021) proposed a novel BiFA-YOLO detector for arbitrary-oriented ship detection in high-resolution SAR images, which incorporates bi-directional feature fusion and angular classification [48]. This approach demonstrates robust detection capabilities with improved precision and recall rates compared to other deep learning methods, offering potential applications in naval surveillance and maritime security.

These domain-specific adaptations highlight the flexibility of YOLO architectures in addressing diverse multispectral object detection challenges across various industries. By tailoring models to each domain's specific requirements and data characteristics, researchers are continually advancing the field of multispectral object detection for real-world use.

**D. Real-Time Applications**
A recurring theme in the literature is optimizing YOLO models for improved performance and efficiency in real-time multispectral object detection applications. This is particularly important in specific domains, such as search and rescue and military applications where time constraints require rapid detection. Several studies have proposed lightweight variants of YOLO for real-time multispectral applications. Li and Ye developed Edge-YOLO, a lightweight infrared object detection model designed for deployment on edge devices [202]. Their approach significantly reduced computational requirements by 70.3% and increased speed by 44% while maintaining the same detection accuracy as default YOLOv5 [202].

Another aspect of performance optimization is the development of more efficient fusion strategies. Wang et al. introduced GT-YOLO for nearshore infrared ship detection, incorporating improved feature fusion capabilities for small target detection [203]. Wu et al. proposed a lightweight cross-modality feature fusion method for enhancing multispectral object detection speed and accuracy [204]. Research is also being conducted on pruning and quantization techniques to reduce model complexity while maintaining performance. Yang et al. developed a lightweight coal gangue identification and detection system using YOLOv8n and multispectral imaging technology, achieving improved detection accuracy and speed with a reduced model size [205].

**E. Chinese Dominance In The Literature**
The survey reveals a striking dominance of Chinese research in multispectral object detection, particularly in YOLO-based approaches. Of the 200 studies analyzed, 116 (58%) originated from China, with Chinese researchers leading 65 out of 83 studies (78%) focused on modifying YOLO algorithms for multispectral applications. This prevalence indicates a significant investment by Chinese institutions in advancing real-time multispectral object detection technologies. Furthermore, Chinese publications in this field achieve an average impact factor of 4.45, slightly higher than the 4.36 average for non-Chinese authors. This marginal difference suggests that while Chinese research is prolific, it also maintains a competitive level of quality and influence in the global scientific community.

A multifaceted approach is necessary to address the disparity in multispectral object detection research in the United States. Increased funding allocation to research institutions and universities, specifically for multispectral imaging and object detection, is crucial. This should be coupled with fostering collaborative initiatives between academia, industry, and government agencies to accelerate innovation and practical applications. Enhancing educational programs in computer vision, multispectral imaging, and related fields will also cultivate the necessary expertise in this domain. Additionally, encouraging cross-disciplinary research can drive innovative applications, particularly collaboration between experts in computer vision, sensor technology, and domain-specific fields such as agriculture, infrastructure, and defense. Finally, streamlining processes for translating academic research into commercial and defense applications will ensure that advancements in multispectral object detection quickly transition from theory to practice. By implementing these strategies, the United States can work towards closing the gap in multispectral object detection research and maintaining technological competitiveness.

**F. Challenges And Future Directions**
The literature also discusses several persistent challenges and emerging directions in YOLO multispectral object detection research. One significant challenge is the limited selection of publicly available annotated multispectral datasets for training and evaluation. Many studies rely on custom datasets or adapt existing RGB datasets, which may not fully capture the complexities of multispectral data. Developing comprehensive, publicly available multispectral datasets remains an important area for future work. This challenge is discussed in further detail in the discussion section (Section VIII).

Another challenge is the effective alignment and registration of data from different spectral modalities. Kim et al. addressed this issue by proposing an uncertainty-guided cross-modal learning approach for robust multispectral pedestrian detection [184]. Yuan et al. introduced a cascade alignment-guided transformer to improve RGB-infrared alignment and fusion [185]. These studies highlight the need for robust methods to handle misalignments and discrepancies between spectral inputs.

The literature also points to the growing interest in incorporating transformer architectures into YOLO models for multispectral applications, as seen in the work of Shao et al. and Zhu et al. [206], [169]. Transformers show promise in improving feature fusion across spectral modalities. Transformer architecture is successfully being used to train and run Large Language Models (LLMs), and their use is slowly growing in the computer vision field.

The cross-cutting themes identified in this section highlight the evolving nature of YOLO-based multispectral object detection research. The field continues to advance rapidly, from architectural innovations and domain-specific adaptations to performance optimization and emerging challenges.

**VI. The Evolution Of Yolo In Multispectral Object Detection**

This section will discuss, analyze, and visualize the adaptation of YOLO neural networks for multispectral applications. Deep learning models have revolutionized computer vision, enabling automated analysis and interpretation of multispectral imagery. Various deep learning architectures, such as CNNs, YOLO, Region-based CNNs (R-CNNs), and Support Vector Machines (SVMs), have been used for object detection and classification tasks in multispectral data [207], [208]. CNNs such as YOLO consist of multiple layers and convolutional filters that learn to extract relevant features from the input data [209].

Researchers have proposed various modifications and enhancements to CNNs to improve their performance on multispectral data. These include the use of attention-based fusion networks to extract and fuse data adequately [210]. Incorporation sensor-specific transfer learning methods are also being used to solve problems in traditional supervised methods for training multispectral models [211]. A recent critical development in deep learning for multispectral image object detection is the use of transformer-based models, such as the Vision Transformer (ViT) and MOD-YOLO [169], [212]. Despite being known for their use in large-language models, transformers are beginning to become incorporated into object detection.

Transformer models have shown superior performance compared to CNNs in various computer vision tasks, such as image classification, object detection, and semantic segmentation [169], [206]. The success of transformer models can be attributed to their ability to capture long-range dependencies and information in the data, which is particularly important for multispectral images that have different spatial and spectral resolution [213].

**A. YOLOv5 Adaptations**

YOLOv5 is a one-stage object detection algorithm that utilizes a backbone network for feature extraction, a neck for feature fusion, and a head for prediction (Fig 8A). The backbone consists of multiple convolutional layers (CBS and C3 modules) that downsample the input image and extract features at different scales. Based on the Feature Pyramid Network (FPN) and Path Aggregation Network (PANet), the neck fuses the multi-scale features from the backbone in both top-down and bottom-up paths. This generates feature maps at three different resolutions (P3, P4, P5) for detecting objects of various sizes. Finally, the head uses these fused feature maps to predict the class probabilities, bounding box coordinates, and object confidences. The head consists of convolutional layers that generate the final detection results. Overall, the backbone extracts features, the neck fuses multi-scale information, and the head makes the final predictions in the YOLOv5 architecture.

Of the 37 YOLOv5 multispectral adaptations surveyed, two will be discussed due to their high number of citations and innovative solutions to adapting the YOLOv5 architecture. The first YOLOv5 adaptation selected was the Multispectral Object Detection YOLO model (MOD-YOLO). MOD-YOLO (Fig. 8B) is a lightweight dual-stream network designed for multispectral object detection that consists of a dual-stream feature extraction network and a transformer-based cross-modal spectral multi-scale feature fusion module called Cross Stage Partial CFT (CSP-CFT) [169]. The dual-stream network processes visible and thermal images separately, and the CSP-CFT module effectively fuses the extracted features. Additional improvements include the VoV-GSCSP module in the network head for optimization and the SIoU loss function for enhanced detection accuracy [169].

Unlike the traditional single-stream YOLOv5, MOD-YOLO is better adapted for multispectral object detection due to its dual-stream architecture and CSP-CFT module. The dual-stream architecture enables effective fusion of visible and thermal image features, leveraging the extraction and fusion of critical features. This enhances MOD-YOLO's ability to detect objects in challenging low-visibility scenes, outperforming YOLOv5 by 4.8% mAP [169].

The second most popular YOLOv5 multispectral variant is the Multispectral Object Detection based on Multilevel Feature Fusion and Dual Feature Modulation (named GMD-YOLO), which is a multispectral object detection network designed for low-light environments (Fig. 8C). Like MOD-YOLO, GMD-YOLO has dual-stream architecture consisting of a dual-channel CSPDarknet53 backbone with a Ghost module for feature extraction and a multilevel feature fusion (MLF) module for cross-modal information. GMD-YOLO also has a Dual Feature Modulation (DFM) decoupling head for enhanced object detection of small objects [170].

Unlike the traditional single-stream YOLOv5 architecture, GMD-YOLO's dual-stream architecture processes visible and infrared images separately. The MLF module enables effective multi-scale feature fusion from different modalities. At the same time, the DFM decoupling head provides task-

**A** Default YOLOv5 Architecture

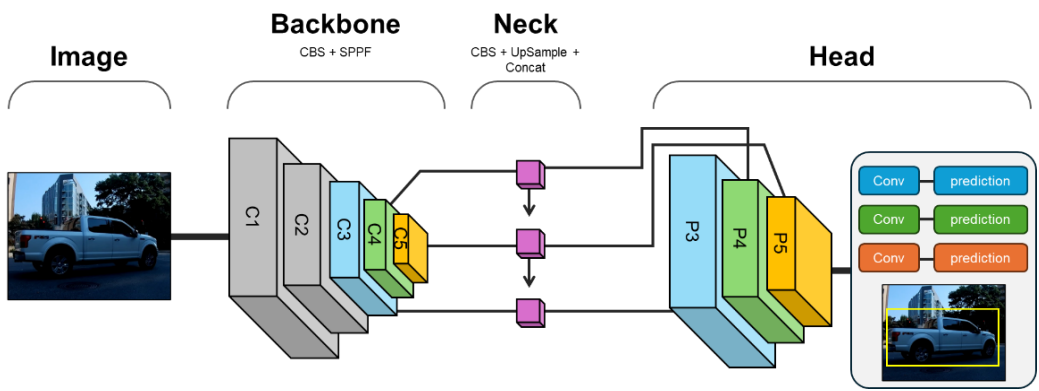

**B** MOD-YOLO(v5) Multispectral Modifications

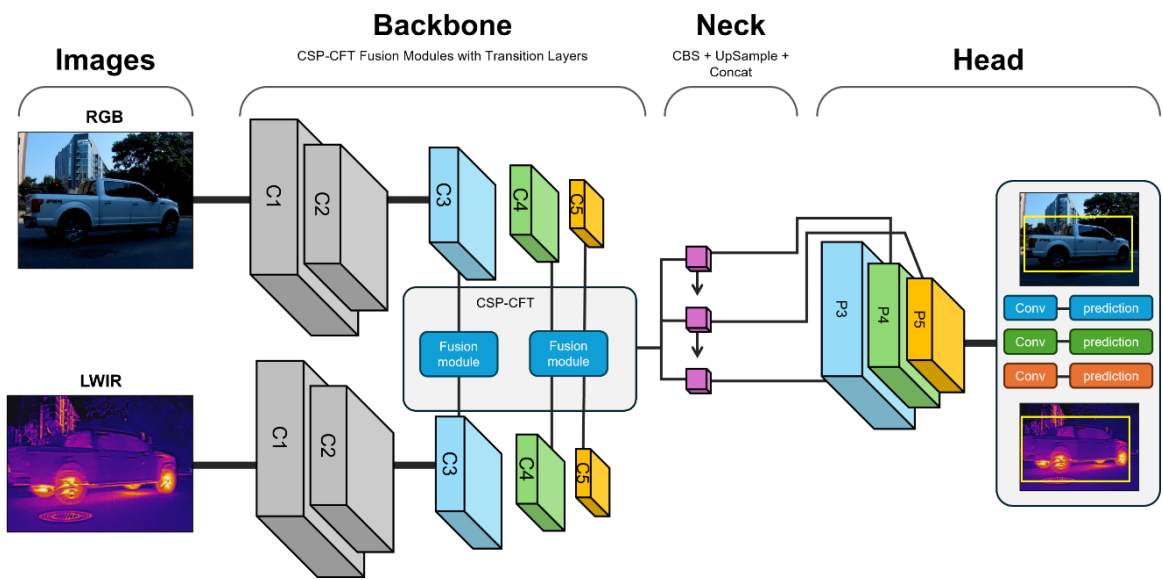

**C** GMD-YOLO(v5) Multispectral Modifications

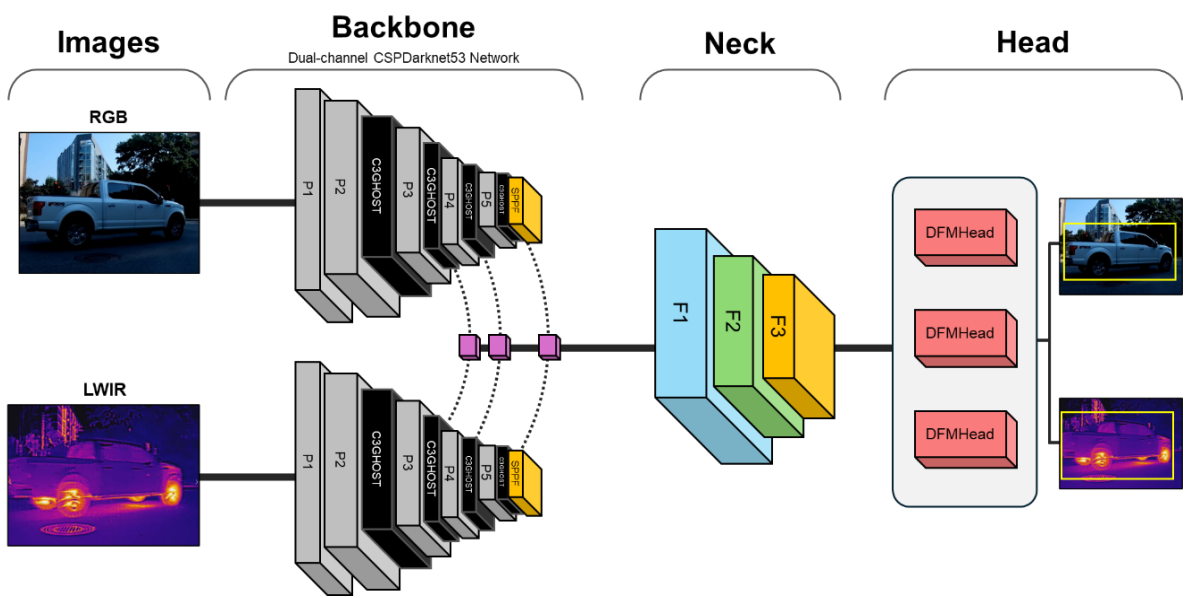

**FIGURE 8**. YOLOv5, MOD-YOLO and GMD-YOLO architectures.

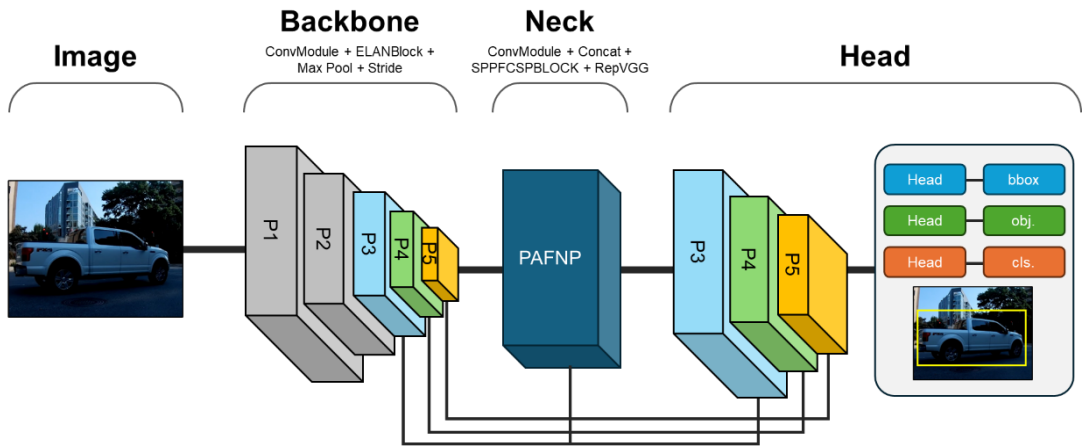

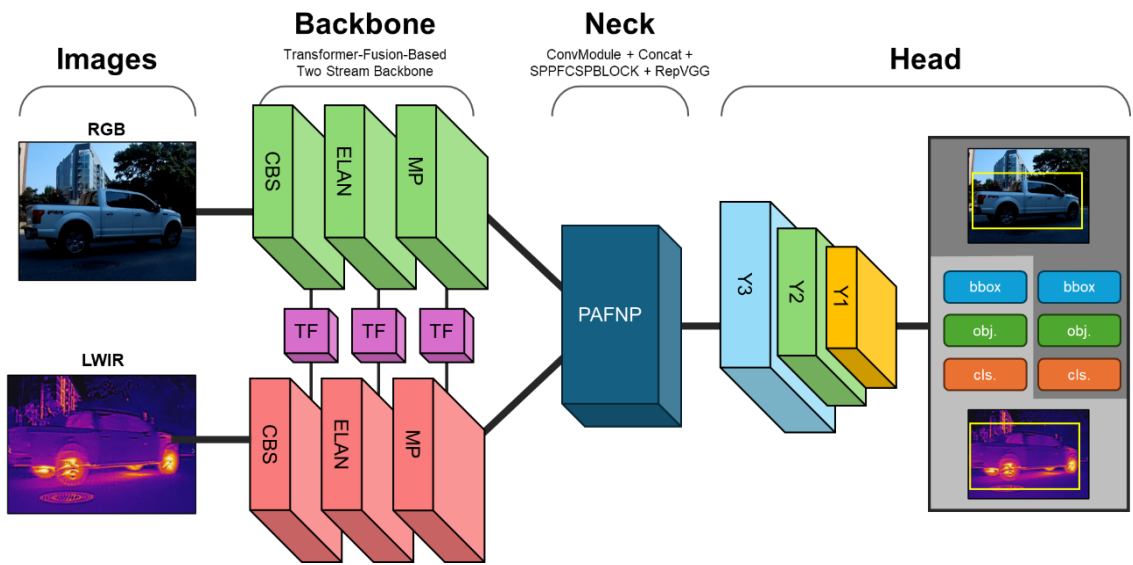

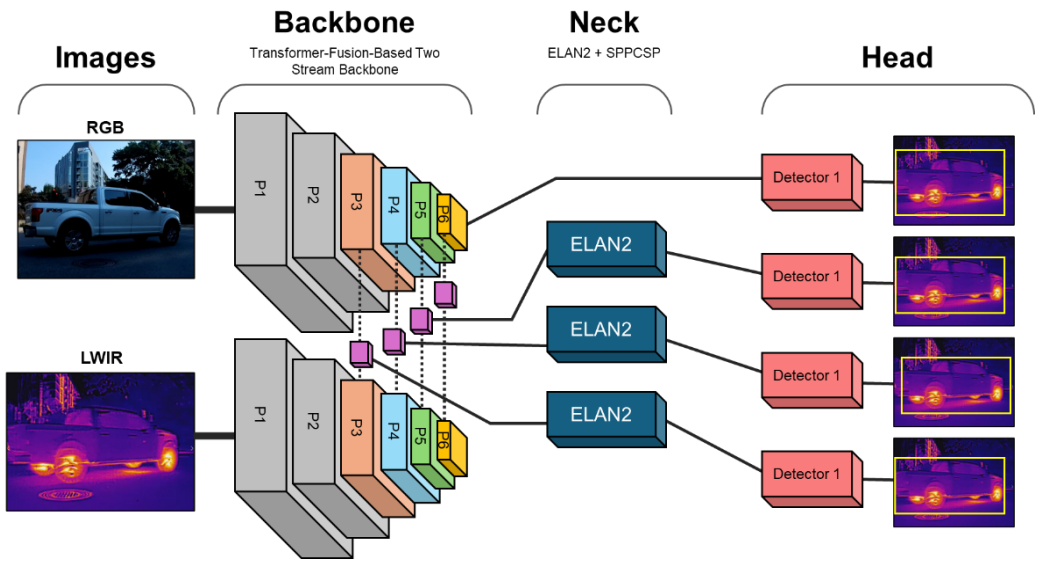

**FIGURE 9.** YOLOv7, TF-YOLO and Dual-YOLO architectures.

specific feature representations, increasing its mAP performance by 3.6% over the default YOLOv5 model [170].

## B. YOLOv7 Adaptations

YOLOv7 was the second most modified YOLO variant. Despite YOLOv7 being the newer of the two models, it is not necessarily better. While YOLOv5 utilizes a focus structure with CSPDarknet53 as the backbone, YOLOv7 incorporates Extended Efficient Layer Aggregation Networks (E-ELAN) and model scaling for concatenation-based models. A study that carried out a comparison between YOLOv5 and YOLOv7 revealed their performance differences. YOLOv5 outperformed YOLOv7 in precision (62.6% vs. 52.8%), mAP@0.5 (55.3% vs. 51.5%), and mAP@0.5:0.95 (34.2% vs. 31.5%), indicating better overall detection accuracy [214]. However, YOLOv7 demonstrates a slightly higher recall value (56.4% vs. 53.4%) [214]. When presented with the custom dataset composed of 9,779 RGB images, YOLOv5 had 4% higher accuracy when compared to YOLOv7 [214]. The pros of YOLOv5 include its lightweight nature, faster inference speed, and higher accuracy, while YOLOv7 offers improved real-time object detection accuracy without increasing inference cost [214].

Figure 9A visualizes the default YOLOv7 architecture. YOLOv7 introduces E-ELAN to enhance the learning capability of the network without disrupting the original gradient path. E-ELAN modifies the architecture in the computational block while keeping the transition layer architecture unchanged [214]. In addition to maintaining the original E-ELAN design, YOLOv7 guides different groups of computational blocks to learn more diverse features. Furthermore, YOLOv7 incorporates a model scaling approach for concatenation-based models, which adjusts specific attributes of the model to generate models of varying scales, catering to different inference speed requirements [21], [62].

The first popular YOLOv7 model modified for infrared applications is Transformers-Fusion-based YOLO (TF-YOLO), a multimodal detection network designed to identify pedestrians under various illumination environments [171]. The novelty of TF-YOLO (Fig. 9B) is its ability to adapt under illumination conditions, a capability that existing multispectral algorithms lack. Like the other modified multispectral YOLO variants, TF-YOLO uses a two-stream backbone network using a transformer-fusion module to integrate critical features between visible and infrared images [171]. The transformer-fusion module allows TF-YOLO to adapt to changing illumination conditions by combining semantic features from high-level layers with high-resolution features from low-level layers. Compared to the traditional YOLOv7 architecture, TF-YOLO incorporates a two-stream backbone for processing visible and infrared images. The primary difference between TF-YOLO and other multispectral YOLO modifications is the use of the transformer-fusion module embedded in the two-stream backbone. By adaptively learning to fuse RGB and infrared features, TF-YOLO can dynamically adapt feature extraction to illumination changes, outperforming YOLOv7-VI by 12.75% mAP and YOLOv7-IR by 8.64% mAP [171].

The next most-cited multispectral YOLOv7 model is Dual-YOLO (Fig. 9C). Dual-YOLO is an infrared object detection network based on YOLOv7 that integrates visible and infrared image features using a dual-branch backbone, attention fusion, and fusion shuffle modules [190]. Dual-YOLO reduces redundant fusion features and enhances complementary characteristics of infrared and visible images. Dual-YOLO introduces attention fusion and fusion shuffle modules to help reduce redundant fusion feature information [190]. The attention fusion module in Dual-YOLO also incorporates Inception and SE modules, thereby enhancing the complementary characteristics of infrared and visible images without increasing parameters. The fusion shuffle module employs dilated convolutions and channel shuffling to increase receptive fields, making infrared and visible features more uniform [190]. Dual-YOLO outperforms YOLOv7-VI by 3.3% mAP and YOLOv7-IR by 5.1% mAP [190].

## VII. Datasets And Evaluation Metrics

The development and evaluation of deep learning models for multispectral object detection rely heavily on the availability of high-quality datasets. Several publicly available datasets have been used in the literature. However, despite the architectural advancements in multispectral object detection models, one of the main limitations is the lack of large annotated datasets to train multispectral models that are on par with RGB datasets. When compared to RGB datasets, there is a significant shortage of multispectral datasets to train and test models [24]. Collecting and labeling such datasets can be time-consuming and expensive, especially for applications requiring expert knowledge [159], [215].

The most common datasets used in this survey were custom datasets (72 studies used custom datasets). The primary benefit of custom datasets is that they can be tailored for specific sensors and object classes. These custom datasets were used for particular detection tasks, such as agricultural crop monitoring [146], [182], [192], [195], [216] and infrastructure inspection [196], [197], [217]. These datasets are collected and curated using multispectral sensors from different platforms, such as multirotor drones and ground systems. Although labor intensive, creating custom datasets is critical for developing domain-specific multispectral models that can handle the unique challenges and requirements of the application. The sensor types that had the highest number of custom datasets in the survey were LWIR and MSI (both had 19 studies), followed by RGB-LWIR (14 studies), and finally, RGB-NIR (8 studies). A primary drawback to curating a custom multispectral dataset for deep learning applications is that it is labor-intensive. A shortfall in many of the custom datasets codified in this survey is that some researchers did not make their data publicly available (44 studies in this survey did not provide a link to access their custom datasets), leading to an inability to reproduce and compare research results.

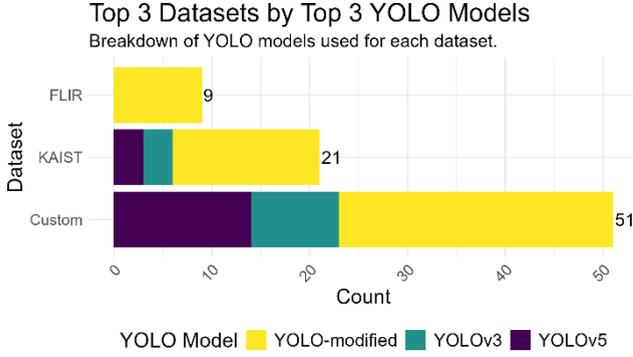

**FIGURE 10.** Top three multispectral datasets used broken down by YOLO model.

The most widely used open-source dataset is the KAIST multispectral pedestrian detection dataset, used by 39 studies in this survey (Fig. 10) [218]. The KAIST dataset is a ground-based dataset of paired RGB-thermal images. Because it is so widely used, the KAIST dataset is considered a benchmark for consistently evaluating multispectral models. Another benefit of the KAIST dataset is its size, consisting of 95,000 color-thermal image pairs manually annotated with three object classes (person, people, and cyclist) accounting for 103,128 total annotations [218]. The next most commonly used dataset in this survey is the ground-based Teledyne FLIR ADAS dataset, used by 22 studies. Although the FLIR dataset has significantly fewer images than the KAIST dataset (only 26,442 images), it has more object class variety [219]. The FLIR dataset has 520,000 bounding boxes across its dataset, consisting primarily of humans, vehicles, street signs, traffic lights, and other everyday objects seen on the road. The FLIR dataset has a near-even split of RGB and thermal images, with a pre-suggested training/validation split to conduct model training out of the box.

The most common air-based dataset used in this survey was the Vehicle Detection in Aerial Imagery (VEDAI), which was used in 9 studies [220]. The VEDAI dataset is derived from satellite images and has multiple vehicle-based object classes, such as car, truck, ship, van, plane, and several other vehicle types. It has four bands consisting of RGB and NIR [220].

## VIII. Discussion

Following our extensive examination of YOLO-based multispectral object detection literature from 2020 to 2024, we present a synthesis of key findings. We return to the three research questions posed at the beginning of this survey, utilizing this discussion to propose recommendations for advancing YOLO-based multispectral object detection.

*1) How have modifications and enhancements to the YOLO architecture impacted its performance and adaptability for multispectral imaging applications compared to default YOLO models?* Adaptations to the YOLO architecture have markedly improved its efficacy and versatility in multispectral imaging applications. The most promising enhancements include the creation of dual-stream architectures, incorporating attention mechanisms and transformer architectures, and developing specialized modules for processing multispectral data. Dual-stream architectures, exemplified by models such as MOD-YOLO and GMD-YOLO, have considerably improved multispectral detection performance over the conventional single-stream CNN approach [169], [170]. These designs process various spectral inputs separately before merging, enabling specialized treatment of each modality. For example, MOD-YOLO surpassed the baseline YOLOv5 by 4.8% mAP, while GMD-YOLO showed a 3.6% mAP improvement compared to the standard YOLOv5 architecture.

Integrating attention mechanisms and transformer architectures has further enhanced YOLO's adaptability to multispectral data. TF-YOLO, which employs a transformer-fusion module, demonstrated dynamic adjustment to varying light conditions, outperforming the standard YOLOv7 by an average of 10.69% mAP [171]. Similarly, Dual-YOLO utilizes attention fusion and fusion shuffle modules to minimize redundant feature information, enhancing the model's ability to focus on relevant spectral characteristics [190]. Specialized modules for multispectral data processing have also been developed to address specific challenges in multispectral object detection. For instance, YOLO-FR introduced a feature reassembly sampling method to enhance small infrared target detection. At the same time, YOLO-SASE combined super-resolution techniques with a novel Self-Adaption Squeeze-and-Excitation module to improve small target detection in complex environments [168], [191]. These modifications have resulted in modified multispectral-YOLO models to consistently outperform traditional YOLO architecture in multispectral object detection tasks.

*2) What sensors, collection platforms, and object classes are growing in use-case applications with YOLO-based multispectral object detection?* The review identified several trends in utilizing various sensors, platforms, and object classes in YOLO-based multispectral object detection applications. There is a notable preference for combining RGB and LWIR sensors, with 62 studies employing this combination. This trend is driven by LWIR's capacity to provide thermal data, offering unique edge features for object detection algorithms and maintaining effectiveness across various visibility conditions. The fusion of RGB and LWIR data has demonstrated promise in enhancing detection performance across diverse lighting conditions. Additionally, NIR sensors are also gaining in popularity, with 14 studies

using this sensor [136]. This is especially true in agricultural applications [182].

Regarding collection platforms, ground-based sensors were the most frequently used, appearing in 127 studies. However, there is a significant trend towards increased use of multirotor drones, with 36 studies employing this platform. The use of drones for multispectral object detection has doubled between 2020 and 2023, reflecting their growing accessibility and versatility. While less numerous, satellite-based platforms (29 studies) show promising results, especially for large-scale monitoring applications. Aircraft were the least utilized platform, appearing in only 12 studies.

The object classes targeted by YOLO-based multispectral object detection span various applications. Human detection, particularly pedestrians, remains a primary focus, especially in autonomous driving and surveillance [218], [219]. However, this trend is likely driven by the limited number of open-source multispectral datasets. Since the KAIST and FLIR datasets have primarily human and vehicle object classes, it is natural that these are the primary object classes being trained and tested by multispectral models. Vehicle detection is another prominent object class, especially in satellite and aerial imagery [220]. In the agricultural domain, there is growing interest in detecting crop diseases, weeds, and specific plant species [44], [45], [192]. Maritime applications have also seen a focus on ship detection [48], [179], [203], [221]. Lastly, infrastructure monitoring using multispectral object detection is also prevalent in the literature [197], [222], [223].

*3) What are the main challenges and future research directions for YOLO-based multispectral object detection?* Despite significant advancements in YOLO-based multispectral object detection, several challenges persist. A previously discussed challenge is the effective alignment and registration of data from different spectral sensors [184], [185]. Misalignments and inconsistencies between spectral inputs, such as parallax and timing issues, can significantly impact detection accuracy. While some studies have proposed solutions, such as the uncertainty-guided cross-modal learning approach by Kim et al. and the cascade alignment-guided transformer introduced by Yuan et al., this remains an area for further research in the field of multispectral object detection [184], [185].

Another challenge is the computational complexities of these models and their constraints in real-time applications [215]. Many studies have proposed lightweight variants, such as Edge-YOLO by Li and Ye, where computational efficiency can increase while maintaining model performance [202]. A previously discussed challenge is the limited availability of large-scale, annotated multispectral datasets for training and evaluating multispectral models. Although limited multispectral datasets allow for the standardized testing of new multispectral YOLO algorithms, they limit model testing for real-world applications. Because of this, many researchers have resorted to creating custom datasets, which is time-consuming and resource intensive.

## IX. Future Research Directions

Based on these challenges, we propose the following future research directions for the field:

*1) Developing adaptive YOLO architectures capable of handling diverse spectral inputs that do not require extensive architectural modifications*: As more systems become automated using computer vision, it is imperative to design lightweight deployable models capable of adapting to complex environments with minimal to no human input [224]. Developing adaptive YOLO architectures capable of handling diverse spectral inputs without requiring extensive architectural modifications is a crucial step toward more versatile and efficient multispectral object detection systems. The current landscape of YOLO-based multispectral object detection often relies on specialized architecture tailored to specific spectral combinations. For instance, models like MOD-YOLO and GMD-YOLO employ dual-stream architectures to process RGB and LWIR inputs separately before fusion. While effective, these approaches require significant architectural changes to accommodate different spectral inputs, limiting their flexibility and necessitating extensive retraining for new spectral combinations.

To address this limitation, a proposed solution is to build a modular YOLO architecture with a dynamic input layer capable of adapting to various spectral inputs. This architecture could employ a bank of spectral-specific preprocessing modules, each designed to handle a particular spectral band or range. These modules would transform the input from each spectral band into a standardized feature representation, which could then be fed into a common YOLO backbone.

The key to this approach is to design a flexible fusion mechanism that can dynamically combine features from different spectral inputs. Drawing inspiration from the attention mechanisms used in TF-YOLO, an adaptive fusion model can be implemented that learns to weigh the contributions of different spectral inputs based on their relevance to the detection task. This would allow the model to effectively utilize information from any combination of available spectral bands without requiring architectural redesign.

To enhance the model's ability to handle diverse inputs, a Neural Architecture Search (NAS) can be implemented to automatically discover optimal subnetwork configurations for different spectral combinations (Similar to YOLO-NAS) [116]. This approach could lead to more efficient and effective processing of varied spectral inputs while maintaining a consistent overall architecture. The development of adaptive YOLO architectures for diverse spectral inputs represents a significant step towards more universal and practical multispectral object detection systems. By reducing the need for extensive and continuous architectural modifications and retraining for different spectral combinations, these models could significantly enhance the applicability of YOLO-based approaches across various domains and sensing platforms. However, a potential challenge in developing such adaptive architecture is maintaining performance across a wide range of spectral inputs without increasing model complexity and computational requirements.

2) *Explore methods to generate large synthetic multispectral datasets with a more available selection of object classes*: When discussing mAP performance, most research papers use the limited publicly available multispectral dataset (such as KAIST and FLIR) to evaluate performance. However, this can lead to deceptive results. For example, YOLOv8 outperforms YOLOv5 on publicly available datasets. However, the YOLOv8 performance decreases and sometimes falls behind YOLOv5 when tested against a custom dataset [224]. To advance the field of multispectral object detection, new implementations must be adopted to generate large synthetic datasets. One such method is the use of Generative Adversarial Networks (GAN). GAN can generate realistic datasets based on original images and has been proven to increase model performance while significantly decreasing the resources required to collect and label data [225], [226]. One such study used GAN to increase the precision and recall of YOLOv5 by over 10% [227].

A method that can be added to generate accurate synthetic datasets that mimic the real world is synthetic data generated by physics-based simulation modeling. Such an approach can generate synthetic datasets that match different environments or applications [228]. For example, a synthetic dataset for different illumination conditions can be generated using GAD and physics-based simulation modeling. Such data diversity would allow for better-performing and more resilient multispectral YOLO algorithms.

When generating synthetic datasets for specific applications, domain randomization techniques can also be used to create realistic data. Domain randomization techniques can generate realistic synthetic data that helps to shrink the reality gap, thereby increasing model performance [229], [230]. This technique is highly beneficial for training multispectral models since variables such as lighting, backgrounds, and object positions can be manipulated [231]. Multi-modal translation techniques can also be used to further augment multispectral data and increase YOLO model performance [232]. These techniques can be applied to large-scale RGB datasets that cover a wide range of object categories to expand the selection of object classes available in multispectral datasets. For example, the COCO dataset, which includes 80 object classes, could serve as a base for generating a multispectral equivalent dataset.

A combination of these approaches can yield robust and diverse synthetic datasets. Furthermore, physics-based simulation could provide a strong foundation for generating realistic spectral signatures, while GANs and domain randomization techniques could introduce the necessary variability to create diverse training datasets. The critical challenge will be to ensure that the generated data accurately represents the complex interactions between different spectral bands and the real-world environment.

Developing standardized evaluation metrics and validation techniques for synthetic multispectral datasets is crucial. This would help ensure that models trained on synthetic data transfer well to real-world scenarios. Collaboration between computer vision researchers and domain experts in spectroscopy and remote sensing will be essential in creating high-quality, physically accurate synthetic datasets.

3) *Advance multispectral YOLO transfer learning techniques to address dataset scarcity*: Advancing multispectral YOLO transfer learning techniques to address dataset scarcity is an important research direction for YOLO-based multispectral object detection. With limitedly available annotated multispectral datasets already identified as an issue, transfer learning offers a promising approach to mitigate this issue. Transfer learning can be used to leverage knowledge gained from models trained on more abundant data sources, such as RGB images, to improve the performance of multispectral models.

One strategy for multispectral YOLO transfer learning is to pre-train models on large RGB datasets and then fine-tune them on smaller multispectral models. By transferring the weights learned from RGB data, the multispectral model can benefit from general feature representations before adapting to the specific characteristics of multispectral data.

Incorporating attention mechanisms and transformer architectures into the transfer learning process could also yield significant improvements. The success of models like TF-YOLO suggests that similar approaches could enhance knowledge transfer between different spectral domains. These architectures could learn more robust and adaptable feature representations that generalize well across different spectral inputs. Another method for transfer learning is the development of modular architectures that can be pre-trained on individual spectral bands and combined for multispectral tasks. This approach would allow for more flexible and efficient transfer learning, as modules could be selectively fine-tuned or replaced based on the specific requirements of the multispectral task. Additionally, exploring meta-learning techniques could enable YOLO models to learn how to adapt quickly to new spectral domains with limited data, potentially addressing the dataset scarcity issue more directly [233].

Furthermore, developing unsupervised or self-supervised learning techniques for multispectral data can also contribute to addressing dataset scarcity. By leveraging the structures and relationships within multispectral data, models could learn meaningful representations without relying heavily on annotated datasets. These pre-trained representations could be a strong foundation for transfer learning in various multispectral object detection tasks. It is also worth considering the potential of few-shot learning techniques in the context of multispectral YOLO models [234]. Given the success of few-shot learning in other computer vision tasks, adapting these approaches to multispectral object detection could enable models to generalize to new object classes or spectral domains with minimal labeled examples [235].

Advancing multispectral YOLO transfer learning techniques presents a promising path to addressing dataset scarcity. The field can work towards more efficient and effective use of the existing limited multispectral data by building upon existing transfer learning methods, incorporating domain-specific adaptations, and exploring novel architectural and learning paradigms. This, in turn, will contribute to the broader adoption and improved performance of YOLO-based multispectral object detection across various applications and industries.

4) *Advancing fusion research with other sensor types beyond RGB and LWIR*: Exploring fusion with sensor types beyond RGB and LWIR represents a promising frontier in advancing YOLO-based multispectral object detection. While the combination of RGB and LWIR has been extensively studied and proven effective, as evidenced by its prevalence in 62 studies in this review, there is potential for further incorporating data from additional sensor types to enhance detection capabilities across a broader range of applications and environmental conditions.

Given NIR's use in agriculture applications, fusing NIR with RGB shows promise. The NIR spectrum is sensitive to vegetation health and water content, making it valuable for precision agriculture and environmental monitoring. Fusing NIR data with RGB and LWIR could provide a more comprehensive understanding, potentially improving detection accuracy for vegetation-related tasks.

However, a significant gap exists in the literature in that more research is needed to quantify the best sensors to fuse given environmental and temporal factors. More research is needed to determine what sensor combinations are optimal for specific environments and how much fusion should occur between two sensor modalities given external variables, such as illumination, ground temperature, climate, object class, and several other factors [236]. Incorporating a temporal-based fusion approach from multiple sensors could enhance the detection of moving objects and small objects in dynamic environments.

Future research should focus on developing efficient, scalable architectures for multi-sensor fusion and creating synthetic or semi-synthetic datasets incorporating a wide range of sensor types, allowing for extensive fusion testing. As discussed in the previous section, transfer learning techniques can also play a significant role in leveraging knowledge from more common fusion methods to improve the performance of less commonly tested sensor combinations.

X. Conclusion

This comprehensive survey of YOLO-based multispectral object detection from 2020 to 2024 reveals significant advancements in the field, driven by architectural innovations, domain-specific adaptations, and performance optimizations. The review of 200 papers from reputable journals highlights several key trends and challenges in this rapidly evolving domain. The dominance of research from Chinese research institutions in this field is evident, with 58% of the studies originating from China, as well as 65 out of 83 studies on multispectral YOLO modifications originating from China as well. This geographical concentration of research efforts underscores the need for broader international collaboration to diversify perspectives and applications in multispectral object detection.

Despite YOLO-multispectral advancements, several challenges persist. The limited availability of large-scale, annotated multispectral datasets remains a significant bottleneck, with 36% of studies relying on custom datasets. This scarcity hampers reproducibility and standardized evaluation of new algorithms.

Future research directions should focus on developing adaptive YOLO architectures capable of handling diverse spectral inputs without extensive modifications, exploring methods to generate large synthetic multispectral datasets, advancing transfer learning techniques to address dataset scarcity, and investigating fusion strategies with sensor types beyond RGB and LWIR. These efforts will be crucial in advancing the field of multispectral object detection and expanding its applications across various industries and domains.

The real-time multispectral object detection field is poised to play an increasingly vital role in several industries. Addressing the identified challenges and pursuing the recommended research directions will further advance the evolution of these models, unlocking new possibilities in computer vision and object detection across a wide range of domains.

Acknowledgment

JG and EO would like to thank the Geography & Geoinformation Science Department at George Mason University for the continual support of their research.

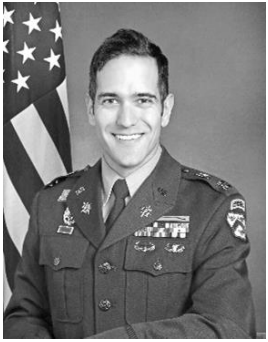
**JAMES E. GALLAGHER** received a B.A from Mercyhurst, an M.A from American Military University, and an MS from George Mason University. He is currently a PhD student at George Mason studying multi-spectral object detection for drone-based applications. He is also an active-duty U.S. Army Major in the Military Intelligence Corps. MAJ Gallagher's research in RGB-LWIR object detection was inspired by his experience using multispectral sensors from UAS in Iraq.

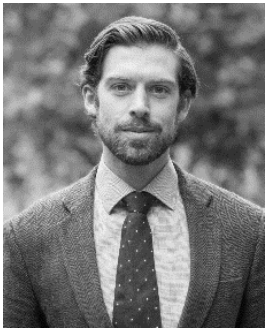
**EDWARD J. OUGHTON** received the M.Phil. and Ph.D. degrees from Clare College, at the University of Cambridge, U.K., in 2010 and 2015, respectively. He later held research positions at both Cambridge and Oxford. He is currently an Assistant Professor in the College of Science at George Mason University, Fairfax, VA, USA, developing open-source research software to analyze digital infrastructure deployment strategies. He received the Pacific Telecommunication Council Young Scholars Award in 2019, Best Paper Award 2019 from the Society of Risk Analysis, and the TPRC48 Charles Benton Early Career Scholar Award 2021.